\documentclass[10pt,twocolumn,letterpaper]{article}

\usepackage[pagenumbers]{wacv} 

\usepackage{graphicx}
\usepackage{amsmath}
\usepackage{amssymb}
\usepackage{booktabs}

\usepackage[pagebackref,breaklinks,colorlinks]{hyperref}

\usepackage[capitalize]{cleveref}
\crefname{section}{Sec.}{Secs.}
\Crefname{section}{Section}{Sections}
\Crefname{table}{Table}{Tables}
\crefname{table}{Tab.}{Tabs.}

\usepackage{algorithmicx}
\usepackage{algorithm,algpseudocode}
\newcommand{\bs}[1]{{\boldsymbol{#1}}}

\usepackage{booktabs}
\usepackage{multicol}
\usepackage{multirow}
\usepackage{listings}
\usepackage{xcolor}
\usepackage{amssymb}
\usepackage{pifont}
\newcommand{\xmark}{\ding{55}}

\usepackage{listings}
\definecolor{commentcolor}{rgb}{0.5,0.5,0.5}
\definecolor{keywordcolor}{rgb}{0,0,1}
\definecolor{stringcolor}{rgb}{0.58,0,0.82}

\def\wacvPaperID{}
\def\confName{}
\def\confYear{}

\begin{document}

\title{Diffusion Cocktail: Mixing Domain-Specific Diffusion Models for \\ Diversified Image Generations}

\author{Haoming Liu \quad Yuanhe Guo \quad Shengjie Wang \quad Hongyi Wen \\
{Shanghai Frontiers Science Center of Artificial Intelligence and Deep Learning, NYU Shanghai} \\
{\tt\small \{haoming.liu, yuanhe.guo, shengjie.wang, hongyi.wen\}@nyu.edu} \\
{\small Project Page: \url{https://MAPS-research.github.io/Ditail}}
}

\twocolumn[{
\renewcommand\twocolumn[1][]{#1}
\maketitle
\centering
\vspace*{-16pt}
\includegraphics[width=0.98\textwidth]{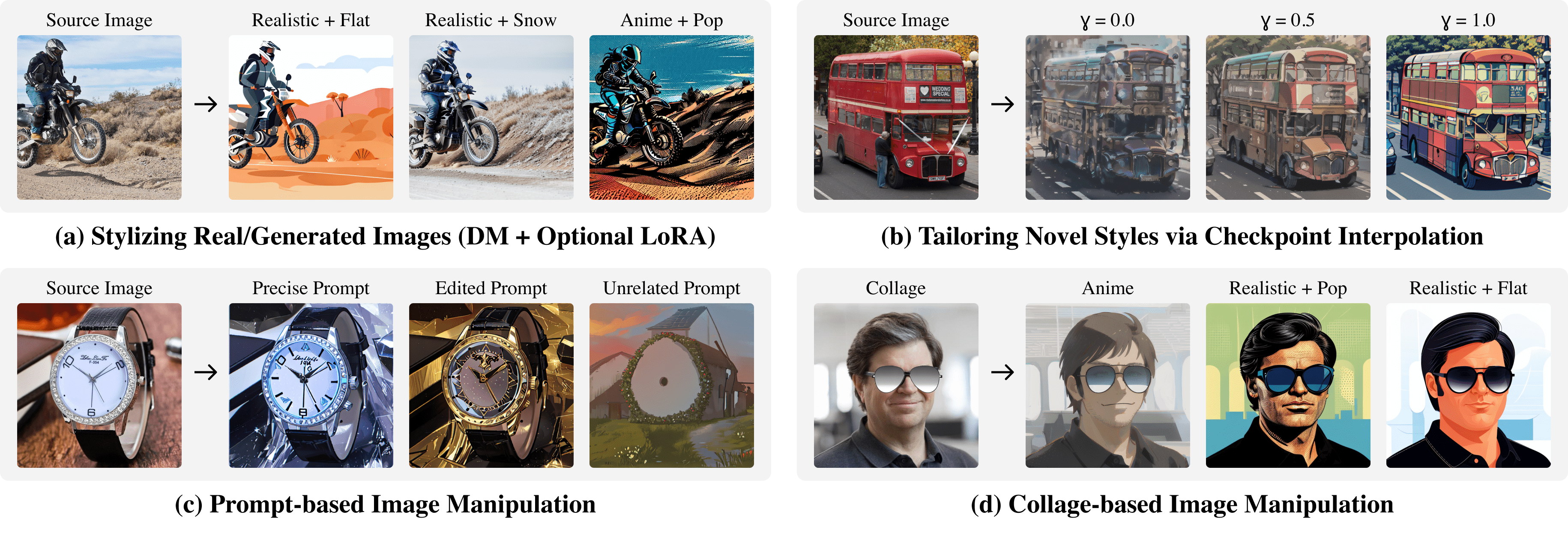}
\vspace*{-12pt}
\captionof{figure}{
Diffusion Cocktail demo use cases.
Ditail primarily focuses on transferring content information between diffusion models. 
We control the generation process of multiple diffusion models by injecting source image structure, resulting in novel images of various contents and styles.
This core idea can be extended to various use cases. Better view with color. Zoom in for best view. 
}
\label{fig:demo}
\vspace*{0.27cm}
}]


\begin{abstract}
Diffusion models, capable of high-quality image generation, receive unparalleled popularity for their ease of extension. Active users have created a massive collection of domain-specific diffusion models by fine-tuning base models on self-collected datasets. Recent work has focused on improving a single diffusion model by uncovering semantic and visual information encoded in various architecture components. However, those methods overlook the vastly available set of fine-tuned diffusion models and, therefore, miss the opportunity to utilize their combined capacity for novel generation. In this work, we propose Diffusion Cocktail (Ditail), a training-free method that transfers style and content information between multiple diffusion models. This allows us to perform diversified generations using a set of diffusion models, resulting in novel images unobtainable by a single model. Ditail also offers fine-grained control of the generation process, which enables flexible manipulations of styles and contents. With these properties, Ditail excels in numerous applications, including style transfer guided by diffusion models, novel-style image generation, and image manipulation via prompts or collage inputs.
\end{abstract}


\section{Introduction}
\label{sec:intro}

Diffusion models (DMs) have shown great success in generating high-quality images~\cite{song2019generative,ho2020denoising,dhariwal2021diffusion}, conditioning on user inputs such as text snippets~\cite{rombach2022high, nichol2021glide, ramesh2022hierarchical, kim2022diffusionclip, ho2021classifierfree}, reference images~\cite{meng2021sdedit} or enriched text~\cite{ge2023expressive}.
A critical feature of the current DMs is the ease of extending and fine-tuning (e.g., ControlNet~\cite{zhang2023adding} or LoRA~\cite{hu2021lora}).
The strong capability in image quality and the easy-to-extend feature of DMs result in a vast community of non-industrial/non-academic users fine-tuning the base DMs to generate images of specific styles or specializing certain characters of interest~\footnote{To date, there are about 8K user-generated DM checkpoints and about 100K LoRAs available on Civitai: \url{https://civitai.com/}.}. This gives us access to a vast collection of fine-tuned DMs with similar model architectures but diverse contents and styles.\looseness=-1

Recent work has focused on studying the conceptual/semantic meaning of various components of a DM to interpret its behaviors for controlling the generation process.
Jeong \etal~\cite{jeong2023training} show that the essential structure information of the generated images is encoded in the self-attention component in the DM's U-net vision transformer.
Tumanyan \etal~\cite{tumanyan2023plug} utilize such an observation to generate new images based on the reference image and additional text prompts while keeping the spatial information.
Kwon and Ye~\cite{kwon2022diffusion} disentangle the content and style information representations in a DM and apply a series of losses to guide an image translation process to manipulate the generated results.
Liew \etal~\cite{liew2022magicmix} mix input semantic concepts by injecting conditioned prompts into the diffusion process and generate images 
of a blended concept given the prompts. \looseness=-1

Even though much work has focused on improving a single DM, rarely have people studied the combinations of diffusion models, ignoring the vast collection of DMs created by everyday users. To the best of our knowledge, we are the first to study image generation using multiple DMs and the transfer of information in between. We propose \textbf{Di}ffusion Cock\textbf{tail} (\textbf{Ditail}), a simple yet effective method that is readily applicable to existing DMs without further training/fine-tuning, thus allowing us to utilize the abundant and fast-growing DM resources efficiently and effectively. \looseness=-1

Ditail enables us to transform the source image (that is not necessarily generated by a DM) to some target domains, using domain-specific DM checkpoints as proxies. By accurately transferring the content information between two DMs during the diffusion process, Ditail can, in theory, handle any-to-any transformation as long as we have DMs fine-tuned on certain domains. In other words, we explore a novel \textit{model-centric style transfer} paradigm, where the target style is set according to a generative model, as opposed to a target style image in the conventional style transfer setting. Given a source image and a target style DM (with optional LoRA), we first extract the content information from DDIM inversion~\cite{song2020denoising} and then apply Ditail to transform the image to the target domain via the target style DM. If no target style model is available, one can mix existing DM and LoRA checkpoints to tailor a novel target style or fine-tune a new checkpoint using a set of style images (essentially defining a target domain based on this set of images). In this way, we can benefit from robust stylization and manipulation capability that no single image could provide.

\begin{figure*}[t!]
  \centering
  \includegraphics[width=\textwidth]{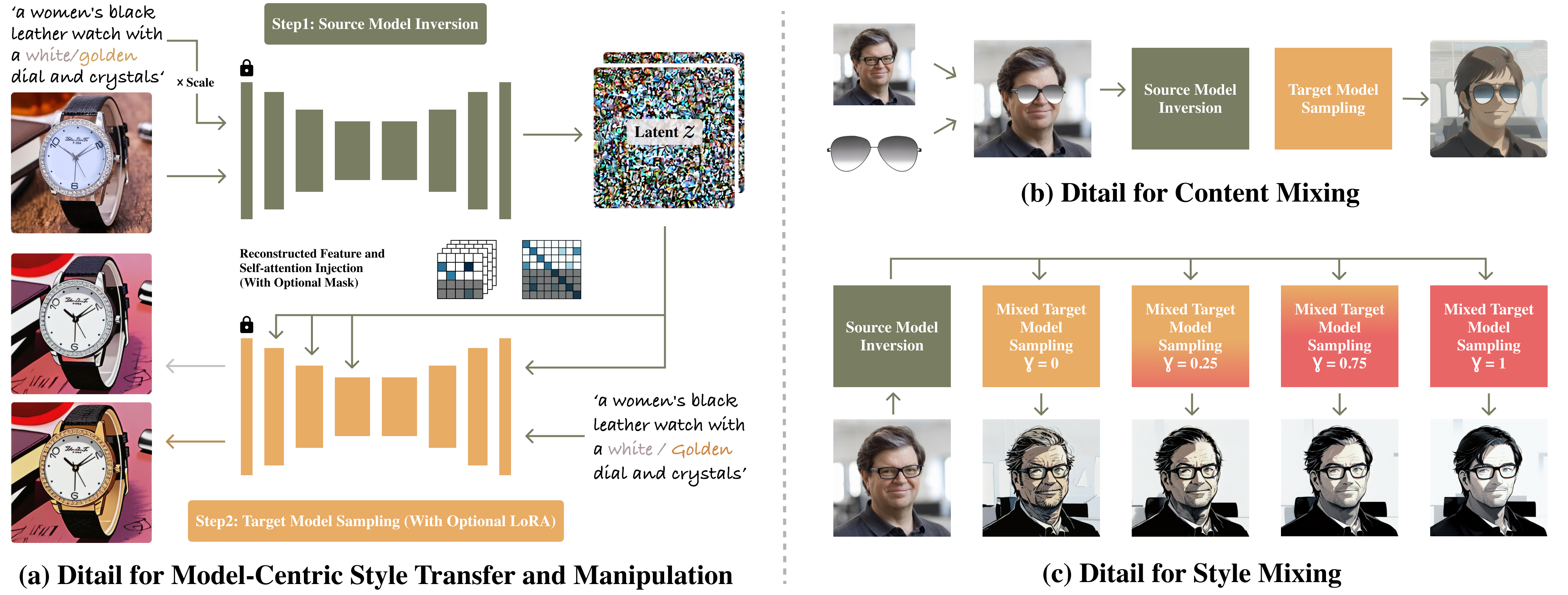}
\caption{
\textbf{(a)} Ditail for model-centric style transfer and content manipulation.
The content image first goes through diffusion inversion, ideally using a DM matching the content image's domain.
The resulting latents are then fed into the style DM for denoising and content injection.
Given various-style DMs, we generate images with their styles while preserving the content image's structures;
\textbf{(b)} Ditail can be naturally extended to transform collages to the same target style/domain; \textbf{(c)} Ditail allows tailoring novel target styles by merging existing style DMs via an interpolation hyper-parameter $\gamma$. Better view with color. Zoom in for best view.
}
  \label{fig:main}
  \vspace{-8pt}
\end{figure*}

Moreover, Ditail is highly extensible to various side applications, such as novel image generation and content manipulation. Given $m$ images generated by $m$ DMs independently with the same textual prompt, one can identify images with desirable contents and styles, and apply Ditail to mix the content and styles for more satisfying results.
Ditail can, in principle, generate $\mathcal{O}(m^2)$ novel images that are not obtainable using a single model from the existing $m$ DMs, significantly increasing the generation flexibility and diversity.
Besides, Ditail can also be used to manipulate the contents by either modifying the textual prompts (e.g., changing the color of the watch in Fig.~\ref{fig:main}(a)) or aligning the appearance of a collage (e.g., adding a pair of sunglasses to an image in Fig.~\ref{fig:main}(b)).
\looseness=-1

Finally, Ditail allows more fine-grained control over the process and uncovers the DM component properties.
We consider both positive and negative prompts in Ditail for content manipulation and further propose a novel condition scaling technique to control the extent of content preservation, thus creating more diversified generation results.

\looseness=-1

\section{Related Work}
\label{sec:related}

\noindent {\bf Diffusion-based style transfer.} 
Wang \etal~\cite{wang2023stylediffusion} first introduce the use of a diffusion model for content extraction and style removal. The diffusion model is fine-tuned to learn disentangled content and style representations. Zhang \etal~\cite{zhang2023inversion} propose to use text conditioning for style transfer with a diffusion model, where they learn the style of a target image through textual inversion. 
Yang \etal~\cite{yang2023zero} utilize a pre-trained diffusion model for the DDIM forward step and use the guidance of gradients computed from designed loss terms that preserve the content structure.
Everaert \etal~\cite{Everaert_2023_ICCV} take multiple style images and computes high-level statistics to learn a style-specific noise distribution. Instead, Ditail preserves the content/spatial features in a source DM and performs transfer style using a target style DM.

\noindent {\bf Image manipulation with a diffusion model.}
Pix2pix-zero~\cite{parmar2023zero} propose an image-to-image translation method that preserves the source image content without manual prompting to guide the translation. It automatically discovers editing directions that reflect desired edits in the text embedding space and retains the source image content by applying a cross-attention guidance loss throughout the diffusion process. Hertz \etal~\cite{hertz2022prompt} identify the importance of cross-attention in controlling the spatial information and its mapping to the text prompt, thus leading to three types of image editing. A few other works have attempted to manipulate image generation with personalized object images as guidance~\cite{ruiz2023dreambooth,ruiz2023hyperdreambooth, gal2022image}. Orthogonally, our work aims to enable personalized manipulation of image styles through the choice of target style DM, a realistic setup with many fine-tuned DMs available.
Tumanyan \etal~\cite{tumanyan2023plug} propose a plug-and-play (PnP) method that edits a given image using a DM with extra prompts to guide the manipulation. PnP first performs diffusion inversion of the given image and extracts features and self-attention maps to record the structure and semantic information. 
Inspired by PnP, Ditail generalizes the concept of PnP to transfer information between two DMs and facilitates downstream tasks.  \looseness=-1

\noindent {\bf Inversion of a diffusion model.}
The inversion problem of a DM aims to find the noise maps as well as the conditioning vector that could generate a given image during the reverse diffusion process. 
Null-text inversion~\cite{mokady2023null} enables intuitive text-based editing of real images with Stable Diffusion models by using an initial DDIM inversion as an anchor, which only tunes the null-text embedding used in classifier-free guidance. 
On the other hand, the inversion of the deterministic DDIM sampling process~\cite{dhariwal2021diffusion} and the inversion of the CLIP~\cite{ramesh2022hierarchical} have been studied to derive a latent noise map for generating specific real images. 
Gal \etal~\cite{gal2022image} propose a textual inversion technique for discovering a new pseudo-word describing the visual concept of a user-provided concept with optimization on only a few images. 

\noindent {\bf Diffusion content-style disentanglement}
Wang \etal~\cite{wang2023stylediffusion} introduce the use of a diffusion model for content extraction and style removal. The diffusion model is then fine-tuned to learn disentangled content and style representations.
Jeong \etal~\cite{jeong2023training} show that the structure information is encoded in the self-attention component in the DM's U-net vision transformer. Based on disentangled representations of content and style, it injects the content information from the h-space (i.e., U-Net's smallest bottleneck) into the style image through diffusion. 
Kwon and Ye~\cite{kwon2022diffusion} focuses on image translation through learning disentangled content and style representations. They propose several training objectives to guide the disentangled representation learning with respect to CLIP and [CLS] representations.

\section{Method}

\subsection{Preliminaries}
Diffusion models~\cite{dhariwal2021diffusion, ho2020denoising, rombach2022high, sohl2015deep} are probabilistic generative models, whose generation consists of the \textit{forward} and the \textit{backward} process. The \textit{forward} process is defined as a Markov chain where each step adds a small amount of Gaussian noise to the latent variables. Formally, for a sequence of time steps $t = 1, \dots, T$, the process is:
\begin{equation}
    \bs{z}_{t} = \sqrt{\alpha_t} \bs{z}_{t-1} + \sqrt{1 - \alpha_t} \bs{\epsilon}_t,
    \label{eq:diffusion_forward}
\end{equation}
where $\bs{z}_t$ is the latent variable at time step $t$, $\{\alpha_t\}$ denotes the variance schedules, $\epsilon_t \sim \mathcal{N}(0, I)$. The backward process aims to reconstruct the latent representation from the noise. It is defined as:
\begin{equation}
    \bs{\hat z}_{t-1} = \frac{1}{\sqrt{\alpha_t}}(\bs{\hat z}_t - \frac{1 - \alpha_t}{\sqrt{1 - \bar{\alpha}_t}} \bs{\hat \epsilon}(\bs{\hat z}_t, t; \theta)),
    \label{eq:diffusion_backward}
\end{equation}
where $\bs{\hat z}_{t-1}$ is the reconstructed latent variable at time $t-1$, $\bar \alpha_t$ is the cumulative product of $\alpha_t$ up to time $t$, $\bs{\hat \epsilon}(\bs{\hat z}_t, t; \theta)$ is the predicted noise from a model parameterized by $\theta$.

Stable Diffusion~\cite{rombach2022high}, a pre-trained text-conditioned Latent Diffusion Model (LDM), has gained attention for its ability to generate high-quality images from textual descriptions. It adopts a pre-trained image autoencoder~\cite{kingma2013vae} to perform the diffusion processes in the latent space with an extra condition on the text embedding $\bs{c}$. It uses a U-Net~\cite{ronneberger2015unet} architecture, where each layer consists of a residual block, a self-attention block, and a cross-attention block. At time step $t$, we denote the output of the U-Net block as $\bs{z}_t$; the residual block outputs as $\{\bs{h}_t^l\}$; the projected self-attention features as $\{\bs{q}_t^l\}, \{\bs{k}_t^l\},\{\bs{v}_t^l\}$, where $l$ denotes layer index.

\subsection{Ditail Content Injection}
Inspired by previous work studying the guidance and translation of DM~\cite{jeong2023training,wang2023stylediffusion,tumanyan2023plug}, we transfer the content information from one DM to another by fusing the U-Net hidden states of the target DM with the ones from the source DM. \looseness=-1

We describe the Ditail algorithm in Alg.~\ref{alg:ditail}.
Ditail takes the DDIM-inverted latents $\{\bs{z}_t^{src}\}_{t=1:T}$ of the source image as input. 
Depending on the use case, Ditail takes positive and negative prompts for extra guidance, and if no prompts are given, we can simply pass in empty strings.
In Line 5-11 of Alg.~\ref{alg:ditail}, Ditail performs the noise sampling procedure.
Particularly, Ditail injects the attention maps $\{\bs{q}_t^{l}\}, \{\bs{k}_t^{l}\}$ and features $\{\bs{h}_t^{l}\}$ of the source image into certain layers of the target DM (we choose feature layer 4 and self-attention layer 4-11 based on previous studies~\cite{tumanyan2023plug}).
The $\mbox{noise-pred}(\cdot)$ function extracts the features to be injected, whereas the $\mbox{noise-pred-with-injection}(\cdot)$ function performs the injection and predicts two noise maps using the positive and negative prompts separately with the target DM.
\looseness=-1

We offer two optional control parameters in Ditail.
The classifier-free guidance $\omega$~\cite{ho2021classifierfree} increases the fidelity of the generated image to the positive prompt.
The mask $\mathcal{M}$ selects the subset of the latent representations for transfer so that only a portion of the content gets injected. Unlike previous work that injects self-attention maps and features from cached tensors, we inject the content information along with the noise sampling of $\bs{z}_t$, where $\{\bs{q}_t^{l}\}, \{\bs{k}_t^{l}\}, \{\bs{h}_t^{l}\}$ can be reconstructed from $\bs{z}_t^{src}$ due to the deterministic nature of DDIM~\cite{song2020denoising}. In other words, Line 7-8 of Alg.~\ref{alg:ditail} can be merged to perform content injection and noise prediction in one operation (see more details in Appendix). Comparing the two, the proposed method greatly reduces the space required to store the injection information (i.e., from 5.7 GB to 4.7 MB in the case of 50 inference steps). One potential caveat here is that we are using different DM checkpoints for inversion and sampling, which may fail to reconstruct the original hidden states; however, our quantitative results over 300 test images and 5 DMs indicate that the performance drops are low enough to be negligible, possibly due to the fact that most DM checkpoints, despite the various styles they exhibit, still share similar weight distributions in the U-Net. Hence, we adopt the proposed content injection paradigm by default to be more memory-friendly and efficient (i.e., it only requires one noise sampling process instead of two).
\looseness=-1

{\small \begin{algorithm}[t]
\caption{$\mbox{Ditail}(\{\bs{z}_t^{src}\}_{t=1:T},P_{pos}, P_{neg}; \theta^{tgt})$}
\begin{algorithmic}[1]
\Statex \textbf{Input:} DM $\theta^{tgt}$, injection latent ${\{\bs{z}_t^{src}\}_{t=1:T}}$ 
\Statex \textbf{Optional Control:} Classifier-free guidance factor $\omega$,  regional injection mask $\mathcal{M}$
\Statex \textbf{Output:} Target image $\mathcal{I}^{tgt}$
\vspace{1mm} \hrule \vspace{1mm}
\State $\bs{e}_{pos}, \bs{e}_{neg} \gets \text{CLIP-enc}(P_{pos}), \text{CLIP-enc}(P_{neg})$
\State $\bs{n} \gets \text{CLIP-enc}(\varnothing)$ 
\State $\Tilde{\bs{c}} \gets \text{concat}(\bs{e}_{pos}, \bs{e}_{neg})$
\State $\bs{z}^{tgt}_T \gets \bs{z}^{src}_T$ 
\For{$t \gets T \ldots 1$}
  \State $\bs{z}_t \gets \text{concat}(\bs{z}^{tgt}_t, \bs{z}^{tgt}_t)$ 
  \State $\{\bs{q}_t^{l}\}, \{\bs{k}_t^{l}\}, \{\bs{h}_t^{l}\} \gets \text{noise-pred}(\bs{z}^{src}_t, \bs{n}, t; \theta^{tgt})$
  \State $\bs{\hat \epsilon}_{t-1}^{pos},\bs{\hat \epsilon}_{t-1}^{neg} \gets \text{noise-pred-with-injection}$
  \Statex\hspace{70pt}$\left(\bs{z}_t,\Tilde{\bs{c}},\mathcal{M}, t; \{\bs{q}_t^{l}\}, \{\bs{k}_t^{l}\}, \{\bs{h}_t^{l}\}; \theta^{tgt}\right)$ 
  \State $\bs{\hat \epsilon}_{t-1} \gets \bs{\hat \epsilon}_{t-1}^{neg} + \omega (\bs{\hat \epsilon}_{t-1}^{pos} - \bs{\hat \epsilon}_{t-1}^{neg})$ 
  \State $\bs{z}^{tgt}_{t-1} \gets \text{DDIM-samp}(\bs{z}^{tgt}_{t}, \bs{\hat \epsilon}_{t-1})$ 
\EndFor
\State \textbf{return} $\text{VAE-dec}(\bs{z}^{tgt}_0)$ 
\end{algorithmic}
\label{alg:ditail}
\end{algorithm}
}

\subsection{Ditail for Model-Centric Style Transfer}
As Ditail excels at injecting content information into the diffusion process while preserving the individual DM's domain specialization, we can also treat Ditail as a method for style transfer between models.
Essentially, we copy the spatial features and self-attention maps from the content DM and inject those into the style DM.
A more precise description for this procedure could be ``content transfer'' instead of style transfer, as the style DM, which received content information from the other DM, generates the target image.

The typical style transfer task takes two images as inputs, and the target is to transfer the style from a style image to the content image while preserving the spatial structure of the content image.
Suppose the content image is generated using a DM of the same architecture as the style DM, and we have access to the features/self-attention maps; then, we can directly apply style transfer using Ditail.
On the contrary, as illustrated in Fig.~\ref{fig:main}(a), given an arbitrary content image (real or generated), we can first extract the structure information using diffusion inversion (i.e., adding reversely predicted noise to the latents using DDIM~\cite{song2020denoising}) and then inject such information using Ditail (Alg.~\ref{alg:styletransfer}).

{\small \begin{algorithm}[t]
\caption{Ditail for Model-Centric Style Transfer and Content Manipulation}\label{alg:main}
\begin{algorithmic}[1]
\Statex \textbf{Input:} Source content image $\mathcal{I}^{src}$, target style DM $\theta^{tgt}$, inversion DM $\theta^{src}$
\Statex \textbf{Optional Control:} Condition scaling factors $\alpha, \beta$
\Statex \textbf{Output:} Target style image $\mathcal{I}^{tgt}$
\vspace{1mm} \hrule \vspace{1mm}
\State Optionally set the prompts $P_{pos}, P_{neg}$ otherwise empty.
\State $\bs{e}_{pos}, \bs{e}_{neg} \gets \text{CLIP-enc}(P_{pos}), \text{CLIP-enc}(P_{neg})$
\State $\bs{c} \gets \alpha \, \bs{e}_{pos} - \beta \, \bs{e}_{neg}$ \Comment{Control the strength of content}
\State $\{\bs{z}_t^{src}\}_{t=1:T} \gets $ DDIM-inv(VAE-enc($\mathcal{I}^{src}), \bs{c}; \theta^{src}$)
\State \textbf{return} $\mbox{Ditail}(\{\bs{z}_t^{src}\}_{t=1:T}, P_{pos}, P_{neg}; \theta^{tgt})$
\end{algorithmic}
\label{alg:styletransfer}
\end{algorithm}}

\begin{figure*}[t!]
  \centering
    \begin{subfigure}{0.58\textwidth}
    \includegraphics[width=\linewidth]{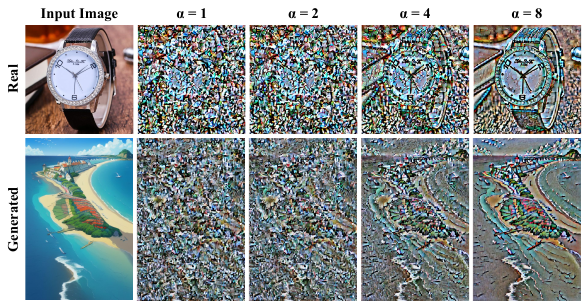}
    \caption{}
    \label{fig:latent}
  \end{subfigure}
    \begin{subfigure}{0.4\textwidth}
    \includegraphics[width=\linewidth]{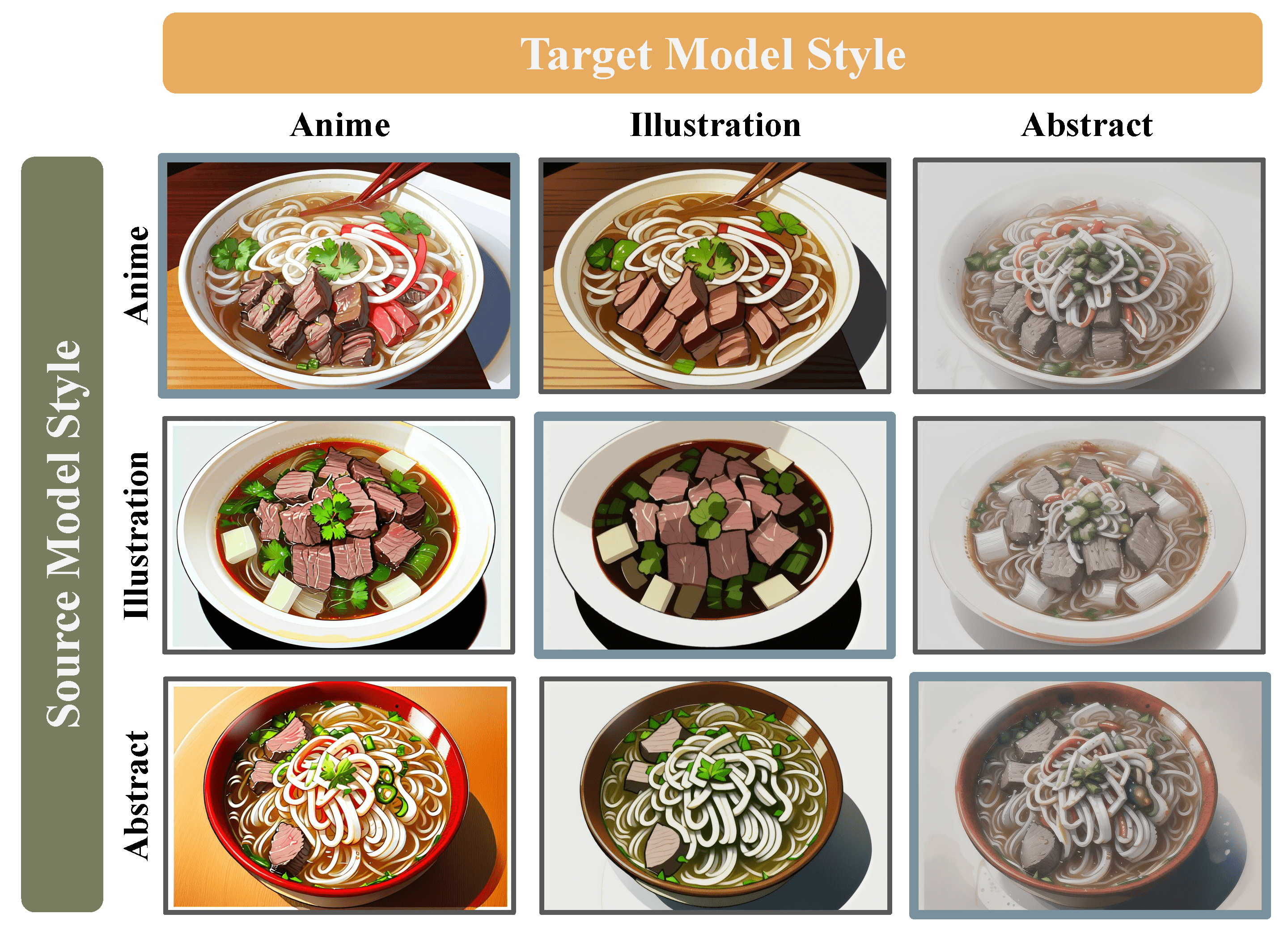}
    \caption{}
    \label{fig:novelgen}
  \end{subfigure}
  \vspace{-8pt}
  \caption{(a) Visualization of noisy latent after DM inversion on real and generated images with different scaling factor $\alpha$. A larger $\alpha$ leads to a stronger structure prior embedded in the noisy latent, resulting in a more structure-preserved starting point in the reverse sampling process; (b) Demonstration of style transfer across different DMs. The diagonal images are the non-modified images generated using each diffusion model with the same textual prompt. Every column corresponds to the images generated by one diffusion model, using contents from other models. Better view with color. Zoom in for best view. \looseness=-1 }
  \vspace{-8pt}
\end{figure*}

In Line 2-3 of Alg.~\ref{alg:styletransfer}, we first generate a guidance condition vector $\bs{c}$, an interpolation between the positive and negative prompt embedding using two hyper-parameters $\alpha, \beta$.
$\alpha$ and $\beta$ are extra knobs to control the strength of the guidance for content injection.
As shown in Fig~\ref{fig:latent}, a larger $\alpha$ value increases the impact of the source image content, mainly generated by interpreting the positive prompt.
Conversely, a larger $\beta$ value does not always yield a significant outcome, as it depends on the semantics of the negative prompt. If the negative prompt is relevant to the content/structure, then $\beta$ will change the resulting content guidance accordingly.
\looseness=-1

We have two extra inputs for applying Ditail on the \textit{model-centric style transfer} task. 
Firstly, the user may provide positive and negative prompts to describe the image to help the inversion process.
In Appendix, we have also tested empty prompts, prompts generated from image captioning, and irrelevant prompts, which verify that both empty and captioning prompts can generate promising results.
Secondly, the user needs to specify a DM for diffusion inversion.
Ideally, a DM matching the content image domain is desirable.
Empirically, the dependence on the inversion model is not strict, as we found that the generation results using different DMs to invert real content images are similar.
Besides, Ditail adopts the invert-and-denoise workflow for the following two reasons: 1) the generating DM or the latent representations are not available in most cases, and 2) we want to adjust the condition scaling parameters $\alpha, \beta$ to strengthen or weaken the content transfer.

We note this is a novel style transfer setting as we use DMs to set the domains/styles of interest.
This allows us to benefit from the vast collection of fine-tuned DMs with drastically different specialization.
If a certain style not included in the available DMs, we can merge existing DM checkpoints to create a novel target style (Sec.~\ref{subsec:tailor}) or fine-tune a new DM model.
Compared to the classic style transfer task, this is equivalent to defining the target style using a set of images.
On the one hand, this gives us more flexibility in defining desirable styles.
On the other hand, the set of images tends to provide more robust and comprehensive style information compared to a single image.

\subsection{Diverse Applications of Ditail}
Based on the core workflow of Ditail (Alg.~\ref{alg:styletransfer}), we can easily extend it to enable diverse applications, such as novel image generation and content manipulation.

\noindent \textbf{Novel Image Generation.} 
Given a pair of DMs $\theta^{(1)}$ and $\theta^{(2)}$, we input the positive and negative prompts to generate two images with each DM independently (denoted as $\mathcal{I}^{(1)}$ and $\mathcal{I}^{(2)}$), where each of them inherits certain contents and styles from their generating DMs. By inverting the images to the noisy latents $\{\bs{z}_t^{(1)}\}_{t=1:T}$ and $\{\bs{z}_t^{(2)}\}_{t=1:T}$ with some condition scalers $(\alpha, \beta)$, we can then apply Ditail to generate two more images, $\mathcal{I}^{(1)\rightarrow(2)}$ and $\mathcal{I}^{(2)\rightarrow(1)}$, each consisting of the structure information of one DM and the style information of the other. Theoretically speaking, given a set of $m$ DMs, we can use Ditail to generate $\mathcal{O}(m^2)$ images over various configurations for condition scaling and target style DMs/LoRAs. 
This dramatically expands the generation capacity as opposed to the $m$ images by running every DM independently, producing images that are not attainable through a single DM.
It also allows the user to have better control over the generation process as well as the integration of more domain-specific DMs.
The user may first use some DMs to generate specific content structure and then transfer the spatial information to other DMs with desirable styles.
Fig.~\ref{fig:novelgen} shows the cross-transfer over three DMs. \looseness=-1

\noindent \textbf{Content Manipulation.}
As shown in Fig.~\ref{fig:demo}, Ditail offers two ways of manipulating the content in the source images: prompt-based and collage-based. Firstly, image content can be directly modified through positive and negative prompts (more examples in Appendix). Secondly, one can create a collage that demonstrates the desirable content structure, such as adding a pair of sunglasses to a human face (Fig.~\ref{fig:main}(b)), then feed it to the Ditail algorithm (Alg.~\ref{alg:styletransfer}) with descriptive prompts and DM selections. Ditail is able to align the appearance of the collage to have higher fidelity. We note that the target domain/style can be identical to the source image, meaning that we can use the same photo-realistic DM for both inversion and sampling stages to get an image of ``A man wearing sunglasses'' in the photo-realistic domain.

\begin{figure*}[t!]
  \centering
\includegraphics[width=\textwidth]{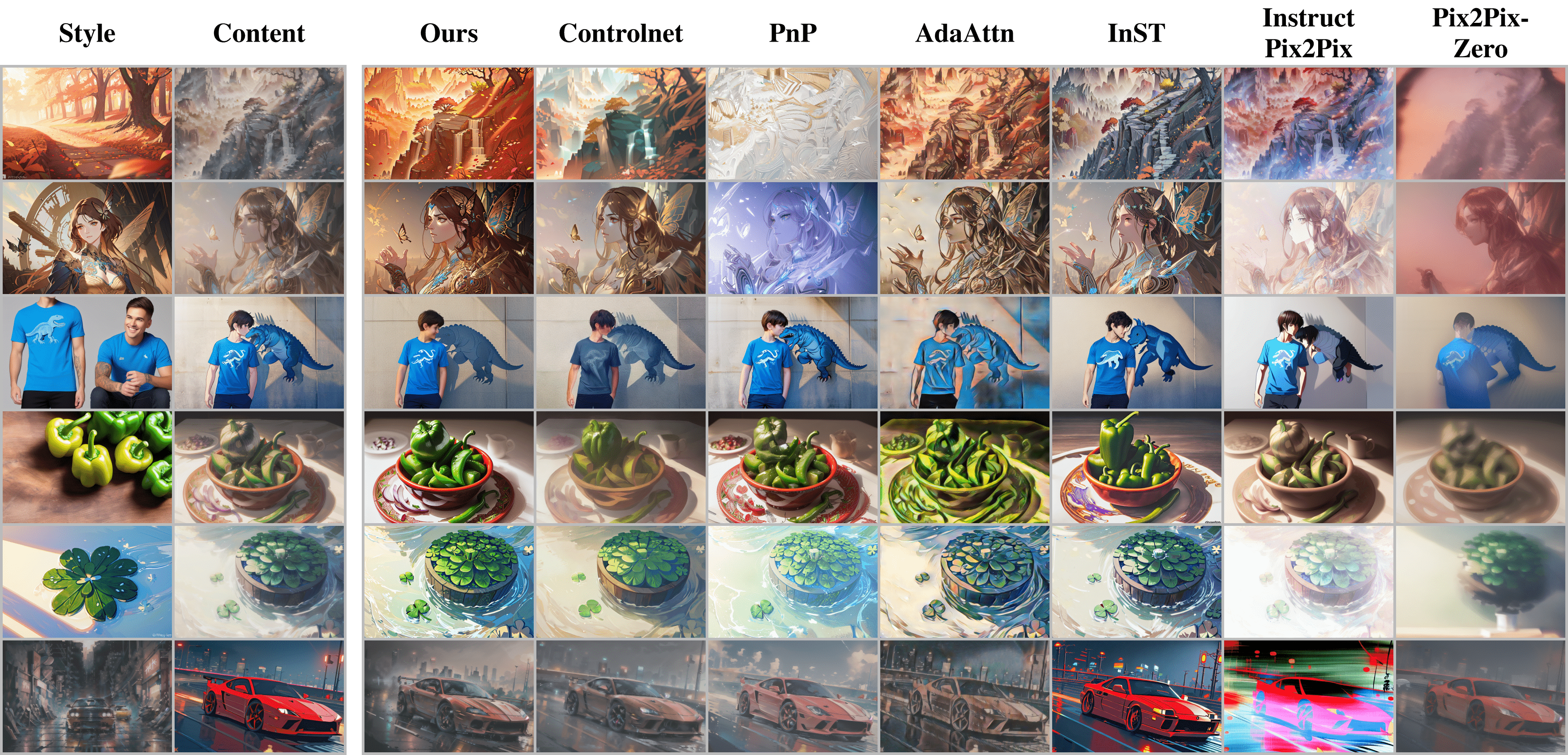}
  \caption{Qualitative comparison of style transfer results on generated images drawn from GEMRec-18K~\cite{guo2023gemrec}. The results of ControlNet~\cite{zhang2023adding} and PnP~\cite{tumanyan2023plug} are obtained with the target DM checkpoint. Better view with color. Zoom in for best view.}
  \label{fig:sota}
  \vspace{-8pt}
\end{figure*}

\subsection{Tailoring Novel Target Domains for Ditail}
\label{subsec:tailor}

Ditail enables flexible target domain tailoring on basis of two key techniques: LoRAs and checkpoint merging. 
LoRA~\cite{hu2021lora} has emerged as a powerful parameter-efficient tuning technique. There is a vast community of fine-tuning LoRAs dedicating to various characters and styles for image generation, which essentially consists of low-rank approximated parameters for the text encoders and attention layers in U-Net. At inference time, such parameters are then scaled and added to a DM base model. For a single layer with weight matrix $W\in\mathcal{R}^{m\times n}$, we fuse the LoRA parameters $B\in \mathcal{R}^{m\times r}$ and $A\in \mathcal{R}^{r\times n}$ via $\hat W = W + \lambda BA$, where $\hat W$ is the fused parameters and $\lambda$ is a weight scaler.
Alternatively, direct checkpoint merging is also a simple yet effective approach for mixing multiple DMs, which essentially computes a weighted sum over all parameters. Formally, to obtain an interpolated style DM $\theta^{'}$ between style DMs $\theta^{(1)}$ and $\theta^{(2)}$, we let $\theta^{'} = \gamma \, \theta^{(1)} + (1-\gamma) \, \theta^{(2)}$, where $\gamma \in [0, 1]$ is a hyper-parameter that controls the interpolation of the two styles. We present some qualitative results of checkpoint interpolation in Appendix.

\section{Experiments}
\label{sec:exp}

\subsection{Results on Generated Images}
\label{subsec:results_generated}

\noindent \textbf{Dataset and Model Source.} Utilizing Ditial for style transfer requires DM checkpoints with various domain specializations. The GEMRec-18K~\cite{guo2023gemrec} dataset contains 18 thousand images generated by 200 text-to-image DMs fine-tuned on Stable Diffusion~\cite{rombach2022high}, where the DM checkpoints were collected from Civitai. Each DM generates 90 images, and the prompts are divided into 12 categories. 
We pick five DM checkpoints with distinctive style specialization (i.e., abstract, anime, fantasy, illustration, and photo-realistic) and perform any-to-any style transfer using the Ditail framework.
We use the images $\{\mathcal{I}^{(1)}_i\}_{i=1:90}$ generated by one source DM $\theta^{(1)}$ as the content image and stylize it with another target DM $\theta^{(2)}$, resulting in $\{\mathcal{I}^{(1)\rightarrow(2)}_i\}_{i=1:90}$.

\begin{figure*}[t!]
  \centering
\includegraphics[width=0.94\textwidth]{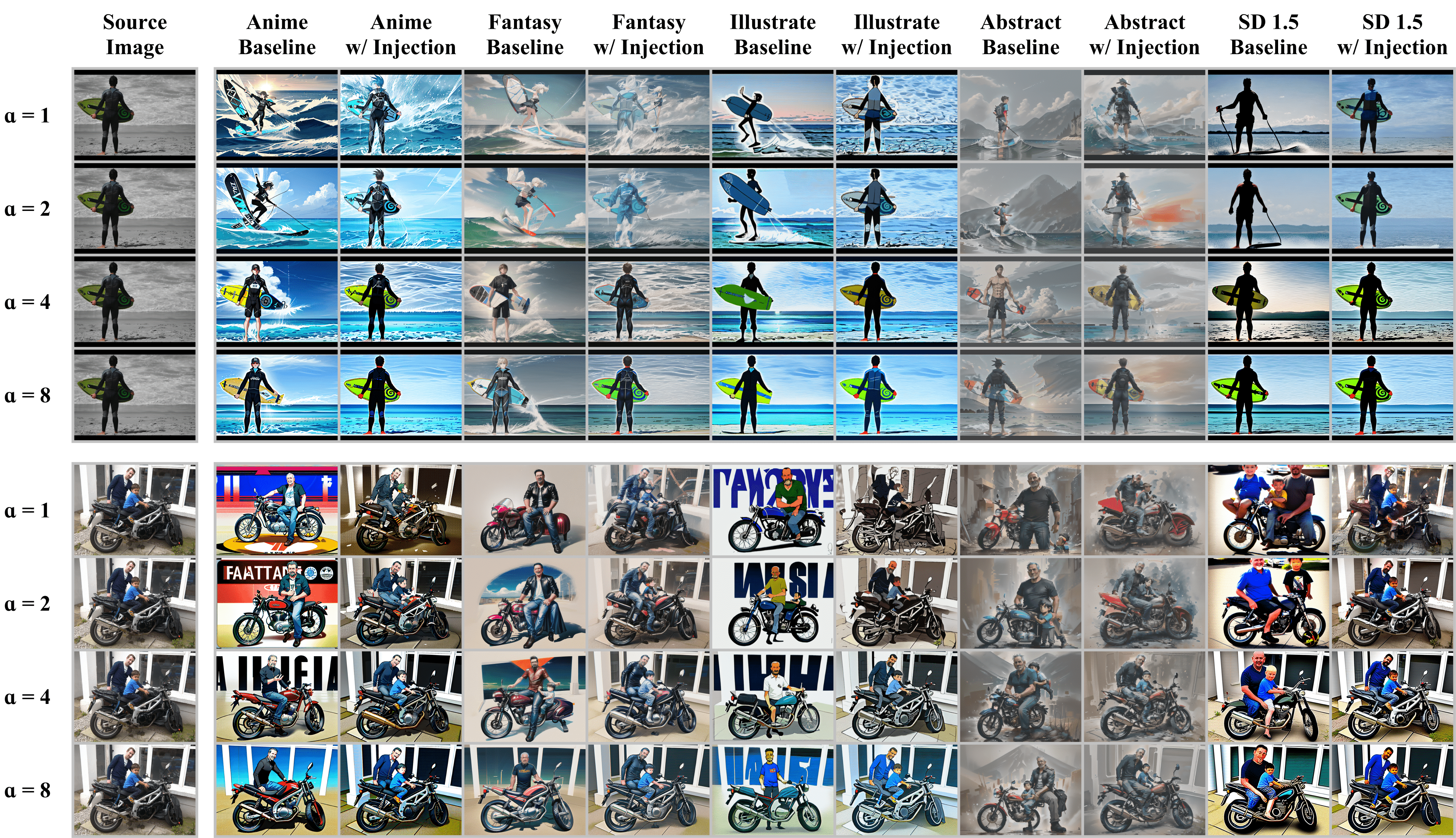}
  \caption{Qualitative results and ablation studies on real images from COCO Captions~\cite{chen2015microsoft}. Better view with color. Zoom in for best view.}
  \label{fig:alpha_beta}
  \vspace{-8pt}
\end{figure*}

\noindent \textbf{Qualitative Results.} We compare the qualitative results with other style transfer~\cite{liu2021adaattn, zhang2023inversion}, image manipulation~\cite{tumanyan2023plug, brooks2023instructpix2pix, parmar2023zero}, and conditional generation~\cite{zhang2023adding} approaches in Fig.~\ref{fig:sota}. The content and style images are generated by $\theta^{(1)}$ and $\theta^{(2)}$ on the same prompt, respectively. Notably, performing style transfer with Ditail does not require a style image due to its model-centric nature, whereas it is a necessity for conventional style transfer approaches. One key observation is that training with one or a few style examples does not guarantee capturing the target style. Such failure will cause a near-unmodified output image, such as the car example for InST~\cite{zhang2023inversion}. AdaAttn~\cite{liu2021adaattn} works in most cases, yet the scattered colors tend to blur the subject in most examples. It is undeniable that the term ``style" has been loosely defined, and most conventional style transfer works tend to specialize in the styles elicited from artistic images. However, we want to expand our scope to a more generalizable setting, which involves domain adaptation capabilities such as ``realistic $\rightarrow$ anime" and ``watercolor $\rightarrow$ illustration", or even the nuance style that a single style image or plain text guidance can hardly capture. In such cases, numerous DM checkpoints with stable styles can serve as a plug-and-play choice and exhibit reasonable performance. As for other approaches, Instruct Pix2Pix~\cite{brooks2023instructpix2pix} and Pix2Pix-Zero~\cite{parmar2023zero} hardly work on this task, whereas ControlNet~\cite{zhang2023adding} and PnP~\cite{tumanyan2023plug} are not robust enough even with the target style DM. More specifically, PnP fails on row 1\&2 and ControlNet fails on row 3\&4. Moreover, Ditail is better on details (e.g., color and lighting) when depicting the same target style, such as row 5\&6.
\looseness=-1

\noindent \textbf{Ablation Study on Content Injection and Condition Scaling.} In this experiment, we study the effects of content injection and various choices of $\alpha,\beta$ values for condition scaling. To quantify the change in image fidelity, structure, and style, we use three metrics, respectively. We use the cosine similarity in CLIP-Score~\cite{hessel2021clipscore, radford2021learning} to measure the compliance between prompt and image; we use the DINO self-similarity loss~\cite{caron2021dino, tumanyan2022splicing} to measure the preservation of structure, which essentially compares the self-attention maps between source and target images; we use Fréchet Inception Distance (FID)~\cite{heusel2017fid} to the images generated by the target model to measure the closeness of style. By the results in Tab.~\ref{tab:generated_metrics}, we observe the following: First, all CLIP scores do not deviate too much from the source image baseline, and the transformed images from most groups move towards the distribution of target images. Second, both enlarging $\alpha$ and introducing $\beta$ leads to a better-preserved structure but pushes the transformed images further away from the target style. Third, the effect of introducing $\beta$ is not monotonic when enlarging $\alpha$, as the negative prompt is generally not related to content or style. Fourth, the best-stylized results are obtained when $\alpha = 2$. Lastly, similar to the conclusions from qualitative results, applying content injection alone already provides a strong baseline for structure preservation.
\looseness=-1

\begin{table}[h!]
\centering
\begin{tabular}{c|c|cccc}
\toprule
Scaling & Injection & CLIP $\uparrow$ & DINO $\downarrow$ & \( \text{FID} \downarrow  \) \\
\midrule
\( \alpha = 1, \beta = 0 \) & \xmark & 0.2586 & 0.0835 & \textbf{175.94} \\
\( \alpha = 2, \beta = 0 \) & \xmark & 0.2591 & 0.0738 & 176.02 \\
\( \alpha = 4, \beta = 0 \) & \xmark & \textbf{0.2606} & 0.0490 & 190.03 \\
\( \alpha = 8, \beta = 0 \) & \xmark & 0.2560 & \textbf{0.0341} & 216.02 \\
\midrule
\( \alpha = 1, \beta = 1 \) & \xmark & 0.2589 & 0.0549 & 189.20 \\
\( \alpha = 2, \beta = 1 \) & \xmark & 0.2579 & 0.0624 & \textbf{183.26} \\
\( \alpha = 4, \beta = 1 \) & \xmark & \textbf{0.2590} & 0.0547 & 185.10 \\
\( \alpha = 8, \beta = 1 \) & \xmark & 0.2562 & \textbf{0.0352} & 213.70 \\
\midrule
\( \alpha = 1, \beta = 0 \) & \checkmark & 0.2546 & 0.0401 & 200.08 \\
\( \alpha = 2, \beta = 0 \) & \checkmark & \textbf{0.2585} & 0.0338 & \textbf{196.38} \\
\( \alpha = 4, \beta = 0 \) & \checkmark & 0.2575 & 0.0261 & 211.11 \\
\( \alpha = 8, \beta = 0 \) & \checkmark & 0.2538 & \textbf{0.0241} & 228.48 \\
\midrule
\( \alpha = 1, \beta = 1 \) & \checkmark & 0.2563 & 0.0280 & 206.82 \\
\( \alpha = 2, \beta = 1 \) & \checkmark & 0.2566 & 0.0304 & \textbf{199.92} \\
\( \alpha = 4, \beta = 1 \) & \checkmark & \textbf{0.2574} & 0.0272 & 204.96 \\
\( \alpha = 8, \beta = 1 \) & \checkmark & 0.2537 & \textbf{0.0245} & 227.55 \\
\bottomrule
\end{tabular}
\vspace{-8pt}
\caption{Quantitative results for model-centric style transfer on generated images from the GEMRec-18K~\cite{guo2023gemrec} dataset. We manually pick 5 models with distinctive styles and conduct style transfer on 90 textual prompts. The metrics are averaged over all 20 possible style transfer combinations. The mean CLIP cosine similarity~\cite{hessel2021clipscore, radford2021learning} on source images is 0.2587, and the mean FID~\cite{heusel2017fid} distance between source and target images is 227.32. The best-performed metric for each group is shown in \textbf{bold}.}
\label{tab:generated_metrics}
\vspace{-16pt}
\end{table}

\subsection{Results on Real Images}
\label{subsec:results_real}

\noindent \textbf{Dataset and Model Source.} We randomly sample 300 images from the COCO Captions~\cite{chen2015microsoft} dataset. We use the same set of style DMs as in the previous experiments.

\noindent \textbf{Qualitative Results.} As shown in Fig.~\ref{fig:alpha_beta}, we perform style transfer on the selected DMs with ablation studies on condition scaling and content injection. We use a unified negative prompt ``worst quality, blurry, NSFW'' by default. 
It can be seen that solely using scaling factors or content injection can preserve the content structure to some extent, and applying content injection, even with reconstructed features and self-attention maps, gives a remarkable structure-preserving performance. Besides, enlarging $\alpha$ leads to better structure preservation effect, which complies with the intuition provided by Fig.~\ref{fig:latent}. In general, applying both techniques together enables us to keep the content unchanged while shifting to the target style seamlessly.

\begin{figure}[t!]
  \centering
    \begin{subfigure}{0.495\linewidth}
    \includegraphics[width=\linewidth]{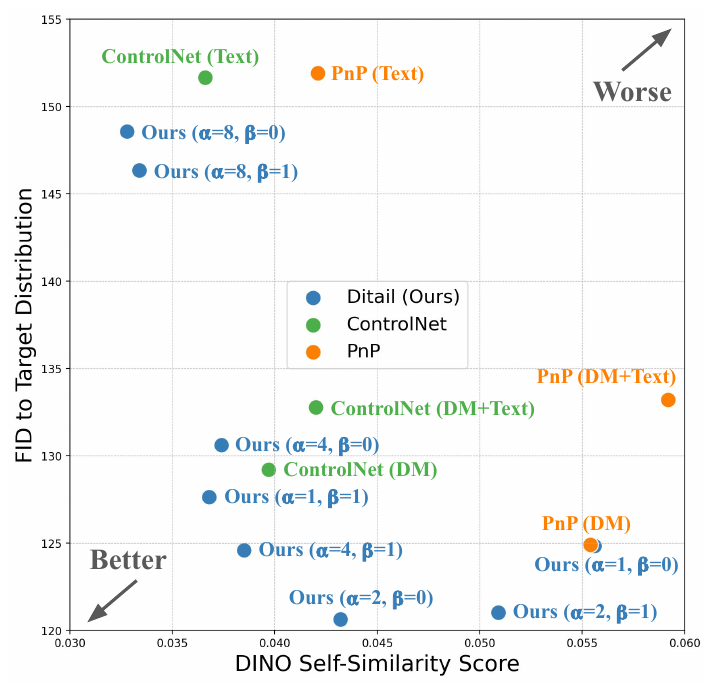}
    \caption{}
    \label{fig:real_heat}
  \end{subfigure}
    \begin{subfigure}{0.48\linewidth}
    \includegraphics[width=\linewidth]{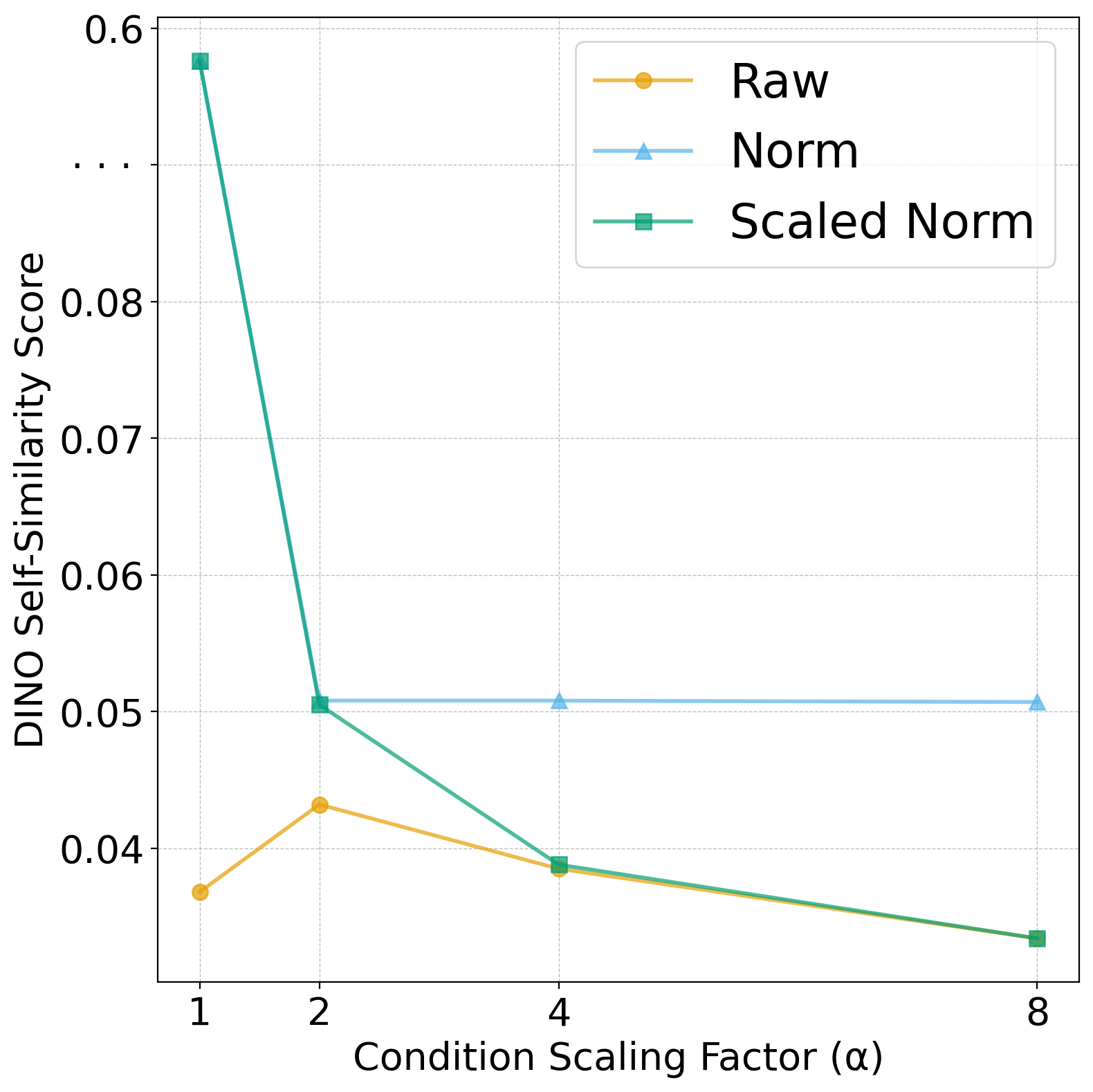}
    \caption{}
    \label{fig:scaling_var}
  \end{subfigure}
  \vspace{-4pt}
  \caption{(a) Comparison of DINO (structure) and FID (style) with PnP~\cite{tumanyan2023plug} and ControlNet~\cite{zhang2023adding}; (b) Comparison of conditional scaling variants on structure preservation; zoom-in for best view. \looseness=-1 }
  \vspace{-8pt}
\end{figure}

\noindent \textbf{Quantitative Comparison with Other Methods.} We make quantitative comparisons of DINO self-similarity scores (structure preservation) and FID scores (style) with the best competitors, PnP~\cite{tumanyan2023plug} and ControlNet~\cite{zhang2023adding}. We consider three variants: prepending style keywords to the textual prompt (on a generic DM), using with the DM checkpoint of target style, and using both of them. As for Ditail, we perform inversion with a photo-realistic DM and then perform noise sampling with the target style DM. We use the condition scaling configurations with the space of $\alpha \in \{1, 2, 4, 8\}$ and $\beta \in \{0, 1\}$. By taking a closer look at Fig.~\ref{fig:real_heat}, we draw the following conclusions: 1) PnP and ControlNet with pre-pending text gives poor stylization performance, suggesting that the nuance styles can be hardly expressed via plain text signals; 2) PnP and ControlNet can exhibit reasonable performance using the target style DM, but may still suffer from robustness issues as shown in Fig.~\ref{fig:sota}, which also demonstrates the superiority of domain-specific DMs compared to generic ones; 3) Ditail enables a flexible trade-off between on structure preservation (DINO) and stylization (FID) via the condition scaling factors $\alpha$ and $\beta$, resulting in better output images in both dimensions.

\noindent \textbf{Why Condition Scaling Works.} To get a better intuition for the condition scaling technique, we compare the proposed version (i.e., direct weighted sum in the Euclidean space, `raw') with two other variants with normalization. Specifically, `norm' normalizes embedding vector $\bs{c}$ after the weighted sum, whereas `scaled norm' rescale the normalized vector by the mean norm. We evaluate the DINO scores with $\alpha \in \{1, 2, 4, 8\}$ and $\beta = 1$. From Fig.~\ref{fig:scaling_var} we can observe that: 1) a wrong condition direction can lead to structure collapse (e.g., $\alpha=1$ in cases with normalization); 2) normalized condition vectors tend to exhibit similar and weaker structure preservation effect; 3) rescaling the normalized condition vectors restores the content to some extent. Hence, the magnitude of condition vectors seem to be more important in maintaining the content consistency. Our hypothesis is that a larger magnitude of the condition vector can interfere the inversion process via the reversely predicted noise, resulting in less random noise being added to the latents and thereby preserving the content. Besides, apart from being a flexible structure preservation technique, and we also empirically found that condition scaling can be helpful in bridging the distribution gap of hidden states from different DM checkpoints.

\section{Limitation and Social Impact}

Despite the diverse side applications empowered by Ditail, it still suffers from a few disadvantages. First, Ditail involves an extra DDIM~\cite{song2020denoising} inversion process, thus resulting in a higher latency compared to conventional text-to-image generations. This can potentially be mitigated via LCMs~\cite{luo2023latent, luo2023lcm}, which substantially reduces the number of inference steps. Second, Ditail may not generalize to mix DMs with distinctive architectures, as the content injection expects hidden states of identical dimensions. Third, Ditail has a strong dependency on checkpoint quality (e.g., fidelity and robustness on the target domain), so the generation results may not be ideal with target DMs of poor quality. Last, Ditail has limited cooperation with human annotations due to training-free nature of the method.

Meanwhile, we also want to emphasize some potential impact elicited by Ditail. First, due to the probabilistic nature of generative models, the images they generate may inadvertently contain potentially unsafe content. Thus, it is essential to adopt content safety checkers in production environment, ensuring that any inappropriate content is filtered before dissemination. Second, as community uploaders fine-tune DMs with private datasets, there are potential copyright concerns regarding the training data and the checkpoint itself. Moreover, the copyright issue also exists for the merged checkpoints with tailored target styles, where the copyright attribution can be tough to determine even with DM fingerprinting techniques. We stay alert to these concerns and maintain ongoing vigilance in monitoring them closely. Last, Ditail's strong capability of diversified image generation can boost the artistic creation of non-academic users on existing DM/LoRA platforms and may potentially lead to business benefits.

\section{Conclusion}

We propose Ditail, a training-free method to transfer content and style information between multiple DMs. 
Ditail allows for image generation by incorporating various DMs, giving users more control and flexibility and benefiting from the extensive collection of DMs.
Ditail applies to style transfer tasks with the target style defined by a DM instead of images. It promises fine-grained manipulations such as textual prompts and condition scaling for controllable image generations. Ditail is highly extensible to diverse application, such as novel-style image generation, prompt-based and collage-based image manipulation. Our future work involves integrating Ditail with other emerging methods and architectures for better efficiency and controllability.

\newpage
\section*{Acknowledgments}
{
    \noindent This work is supported by the Shanghai Frontiers Science Center of Artificial Intelligence and Deep Learning at NYU Shanghai, NYU Shanghai Boost Fund, and NYU High Performance Computing resources and services.
}

{\small
\bibliographystyle{ieee_fullname}
\bibliography{egbib}
}

\newpage
\thispagestyle{empty}
\appendix

\section{Model Checkpoints Information}
\label{sec:model_checkpoints_info}
\begin{figure}[h!]
  \vspace{-8pt}
  \centering
\includegraphics[width=0.94\linewidth]{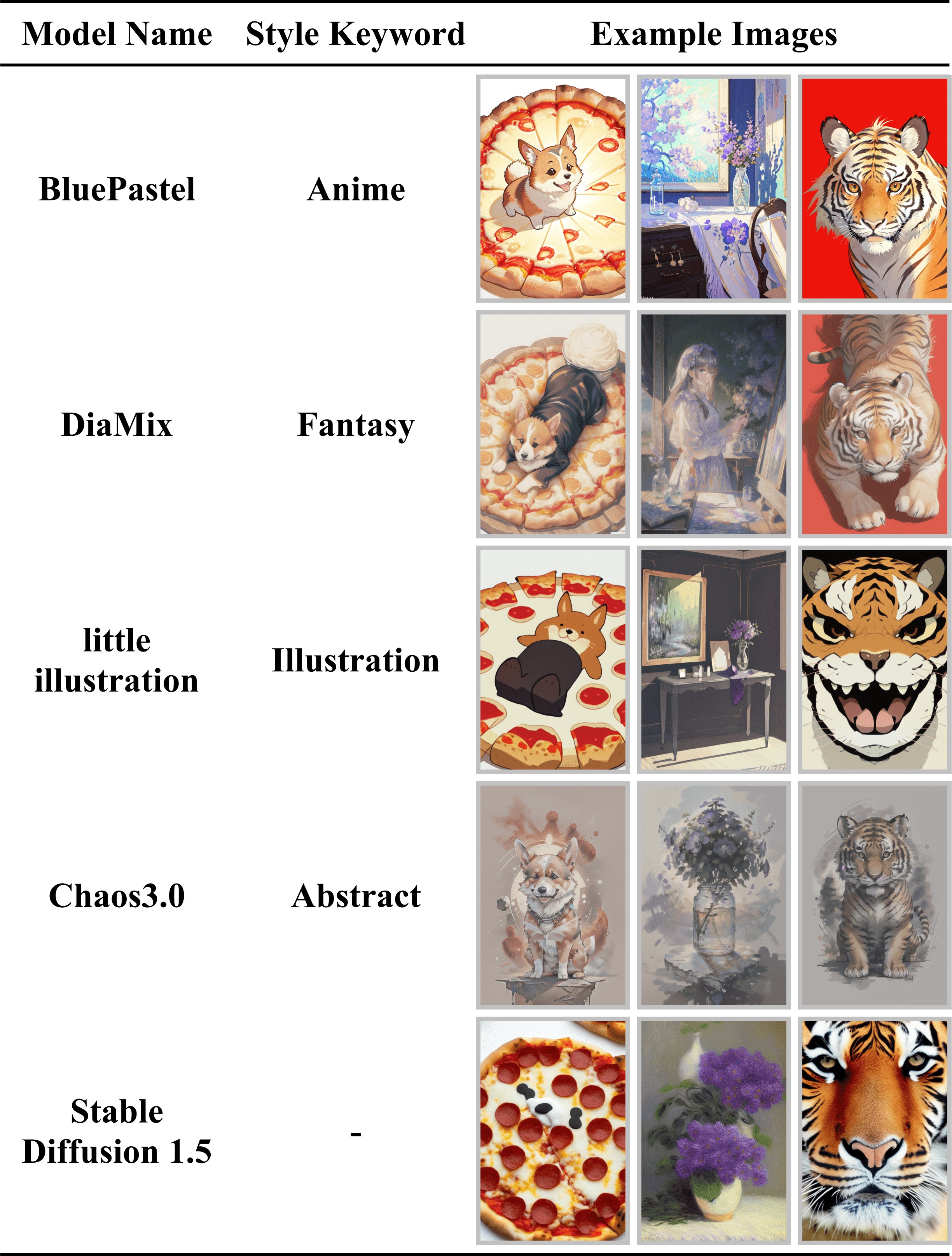}
  \vspace{-4pt}
  \caption{List of selected DM checkpoints from Civitai. Style keyword for each checkpoint are assigned by us. Example images are drawn from GEMRec-18K~\cite{guo2023gemrec}. \looseness=-1}
\label{fig:sm_model_table}
\vspace{-16pt}
\end{figure}

\begin{figure}[h!]
  \centering
\includegraphics[width=0.94\linewidth]{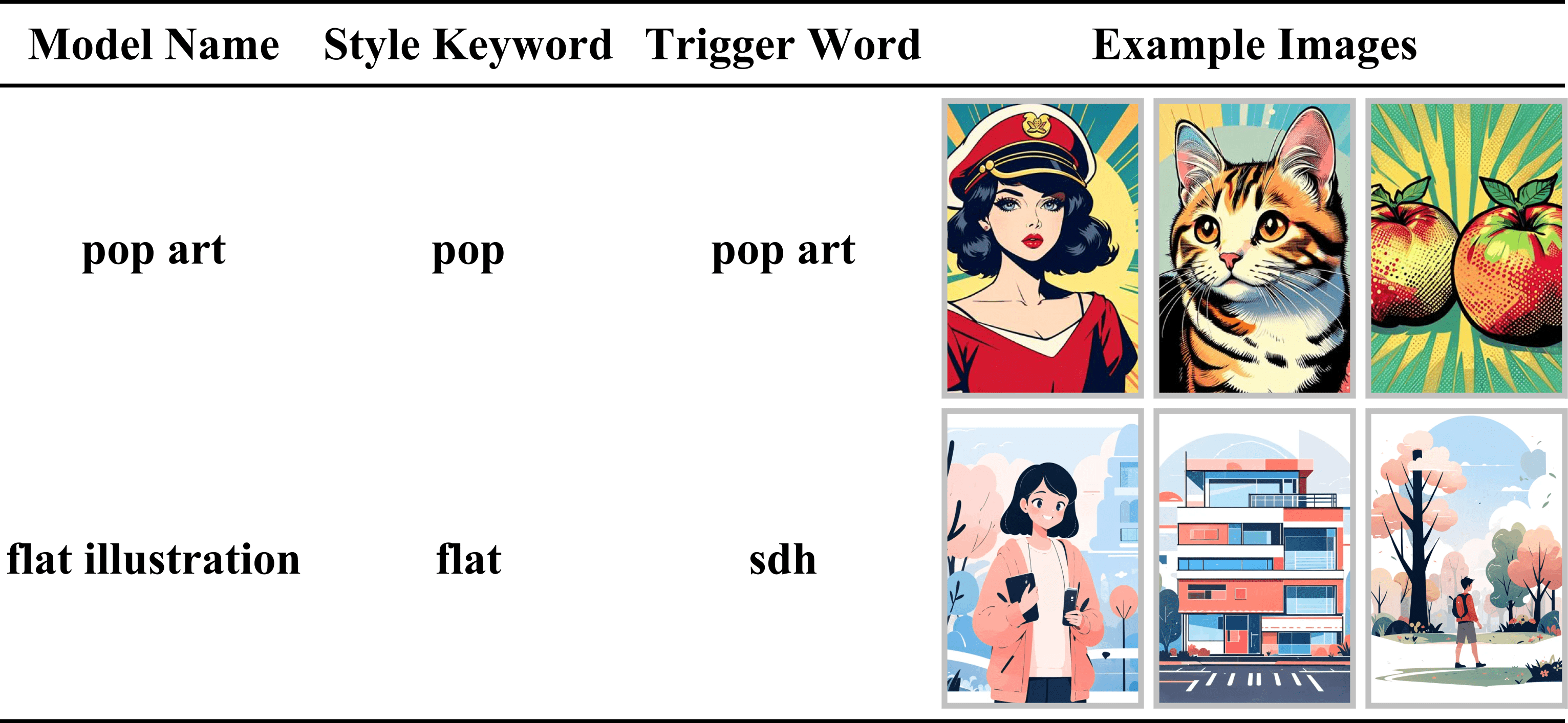}
  \vspace{-4pt}
  \caption{Partial list of selected LoRA checkpoints from civitai and liblib. Style keyword for each checkpoint are assigned by us. Example images are generated by LoRA uploaders. \looseness=-1}
\label{fig:sm_lora_table}
\end{figure}

We manually select 5 DM checkpoints with distinctive style specialization from GEMRec-18K~\cite{guo2023gemrec}, which consists of 200 DM checkpoints from Civitai~\footnote{\url{https://civitai.com}}. Fig.~\ref{fig:sm_model_table} lists the important information about these models, including the style keyword and the tags provided by model uploaders. We also select some LoRA checkpoints from Civitai and Liblib~\footnote{\url{https://www.liblib.ai}} for demonstration, whose information could be found in Fig.~\ref{fig:sm_lora_table}. 

\newpage
\section{Pseudocode for noise-pred-with-injection($\cdot$)}
\label{sec:pseudocode}
\begin{lstlisting}
# Set injection threshold
inference_steps = 50
thresh_res = 0.8 * inference_steps
thresh_attn = 0.5 * inference_steps

# Input:
#   - m: ResNet block to be forwarded
#   - mask: Optional mask for local injection
#   - x: Input hidden states
# Output: Forwarded hidden states
def inj_forward_res(m, mask, x):
    h_src, h_pos, h_neg = m.forward(x).chunk(3)
    h_pos = mask * h_src + (1 - mask) * h_pos
    h_neg = mask * h_src + (1 - mask) * h_neg
    return concat(h_src, h_pos, h_neg)

# Input:
#   - m: Self-attention block to be forwarded
#   - mask: Optional mask for local injection
#   - x: Input hidden states
# Output: Forwarded hidden states
def inj_forward_attn(m, mask, x):
    q, k, v = m.to_q(x), m.to_k(x), m.to_v(x)
    q_src, q_pos, q_neg = q.chunk(3)
    k_src, k_pos, k_neg = k.chunk(3)
    q_pos = mask * q_src + (1 - mask) * q_pos
    q_neg = mask * q_src + (1 - mask) * q_neg
    k_pos = mask * k_src + (1 - mask) * k_pos
    k_neg = mask * k_src + (1 - mask) * k_neg
    q = concat(q_src, q_pos, q_neg)
    k = concat(k_src, k_pos, k_neg)
    return m.attn_forward(q, k, v)

# Input:
#   - unet: U-Net model
#   - x: Input noisy latents
#   - t: Current time step
#   - inj_ids: List of block IDs with injection
# Output: Denoised latent
def noise_pred_with_inj(unet, x, t, inj_ids):
    for i, m in enumerate(unet.blocks()):
        if i in inj_ids:
            if is_res(m) and t >= thresh_res:
                m.forward = inj_forward_res
            if is_attn(m) and t >= thresh_attn:
                m.forward = inj_forward_attn    
    return unet.forward(latents)
\end{lstlisting}

\section{Implementation Details}
We implement Ditail using PyTorch and Diffusers~\footnote{\url{https://huggingface.co/docs/diffusers/index}}. We use Denoising Diffusion Implicit Models (DDIM) scheduler~\cite{song2020denoising} with 50 inference steps for both inversion and sampling. We empirically set $(\alpha, \beta) = (3.0, 0.5)$ by default to balance structure preservation and stylization performance. On a Nvidia A100 GPU, processing 300 images takes $\sim$25 minutes using FP16 precision (i.e., $\sim$5s/img).

\begin{figure*}[t!]
  \centering
\includegraphics[width=\textwidth]{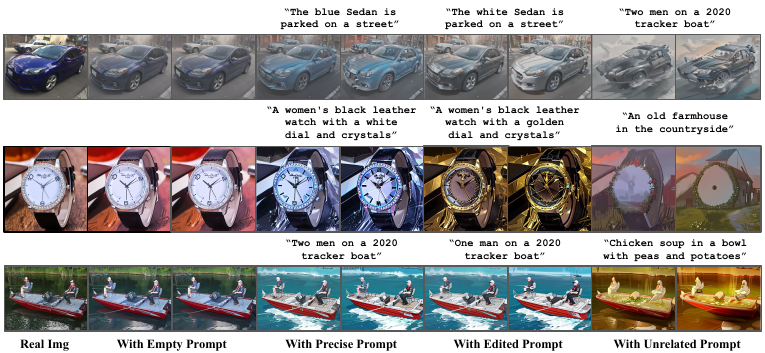}
  \caption{Qualitative results for prompt-based content manipulation. We set the classifier-free guidance factor $\omega = 7.5$ (left) and $15$ (right) for each image pair. The demonstration images are drawn from the LAION-5B~\cite{schuhmann2022laion} dataset with textual prompts generated by the captioner model from BLIP-Large~\cite{li2022blip}. Better view with color. Zoom in for best view.}
  \label{fig:real_edit}
\end{figure*}

\newpage
\section{Prompt-based Image Manipulation}
The qualitative results over various prompt types are shown in Fig.~\ref{fig:real_edit}. We can observe that: 1) an empty prompt exhibits limited editing power compared to the others; 2) a larger CFG scale $\omega$ allows the target models to be more creative; 3) given
that we enforce a strong structure emphasis through condition scaling and content injection, the editing prompts that change the number of objects usually give unsatisfactory results, whereas modifying the semantic class and/or colors of the subjects seems to be easier, even for completely unrelated prompts (e.g., man $\rightarrow$ chicken).

\section{Comparison of Injection Methods}
\begin{figure}[h!]
  \centering
\includegraphics[width=\linewidth]{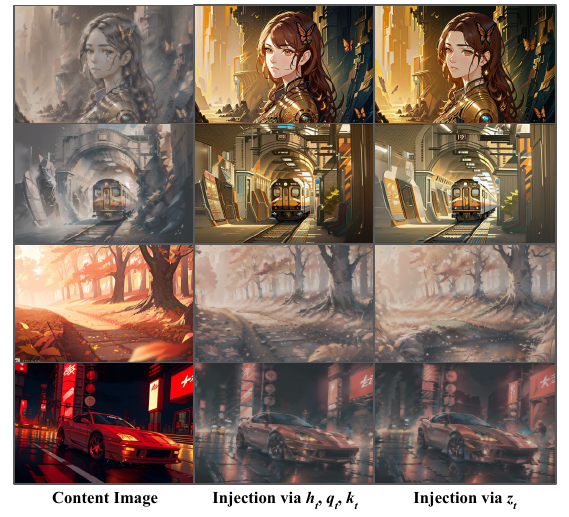}
    \caption{Qualitative comparison of content injection methods. The middle column refers to content injection via cached tensors (i.e., $\{\bs{h}_t^{l}\}, \{\bs{q}_t^{l}\}, \{\bs{k}_t^{l}\}$). The right column refers to our proposed injection via reconstructing from latent $\bs{z}_t^{src}$. \looseness=-1}
    \label{fig:injection}
  \vspace{-4pt}
\end{figure}

Due to the memory burden of storing tensors (5.7 GB/img), we conduct evaluation over 300 randomly sampled (image, target style) pairs. The performance drops on CLIP, DINO, FID are 0.0013, 0.0018, 0.23, respectively, which are low enough to be negligible. Some qualitative results are shown in Fig.~\ref{fig:injection}, and it can be seen that the differences are relatively minor, where the proposed injection method is more memory-friendly (4.7 MB/img) and efficient (i.e., it reconstructs the hidden states in parallel, so it only needs one noise sampling process).

\begin{figure*}[ht!]
  \centering
\includegraphics[width=\textwidth]{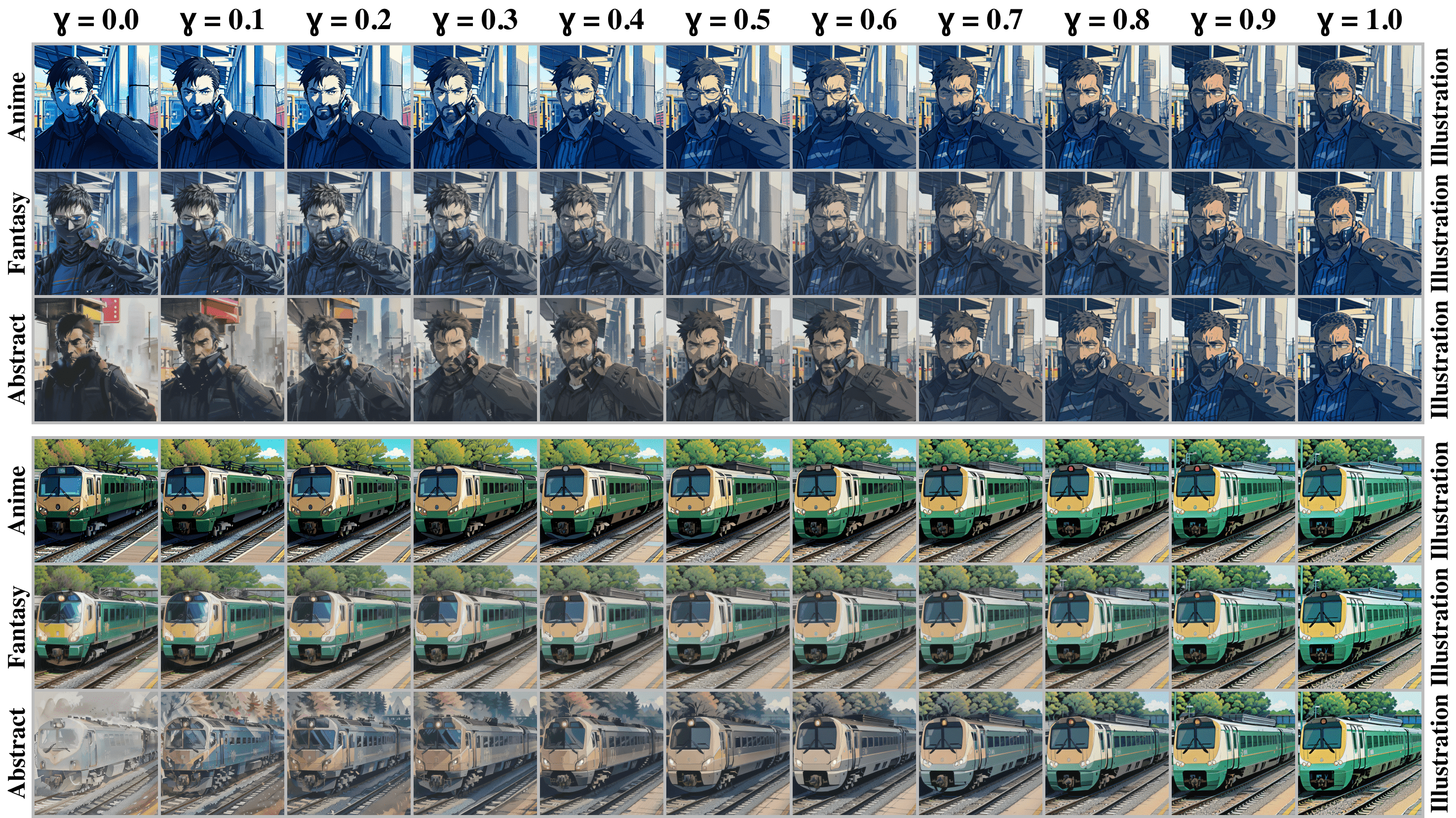}
  \caption{Style mixing via checkpoints interpolation. Better view with color. Zoom in for best view.}
  \label{fig:style_mix}
\end{figure*}

\newpage
$ $ \newpage

\section{Qualitative Results for Style Interpolations}
We present some qualitative results of interpolation between different styles in Fig.~\ref{fig:style_mix}. By adjusting the interpolation factor $\gamma$, we can observe a smooth transition from one style to another. This technique allows us to tailor infinite target styles by merging existing DM checkpoints.

\section{Comparison of Different Inversion Models}
\label{sec:inversion_model}

\begin{figure*}[t!]
  \centering
\includegraphics[width=0.9\textwidth]{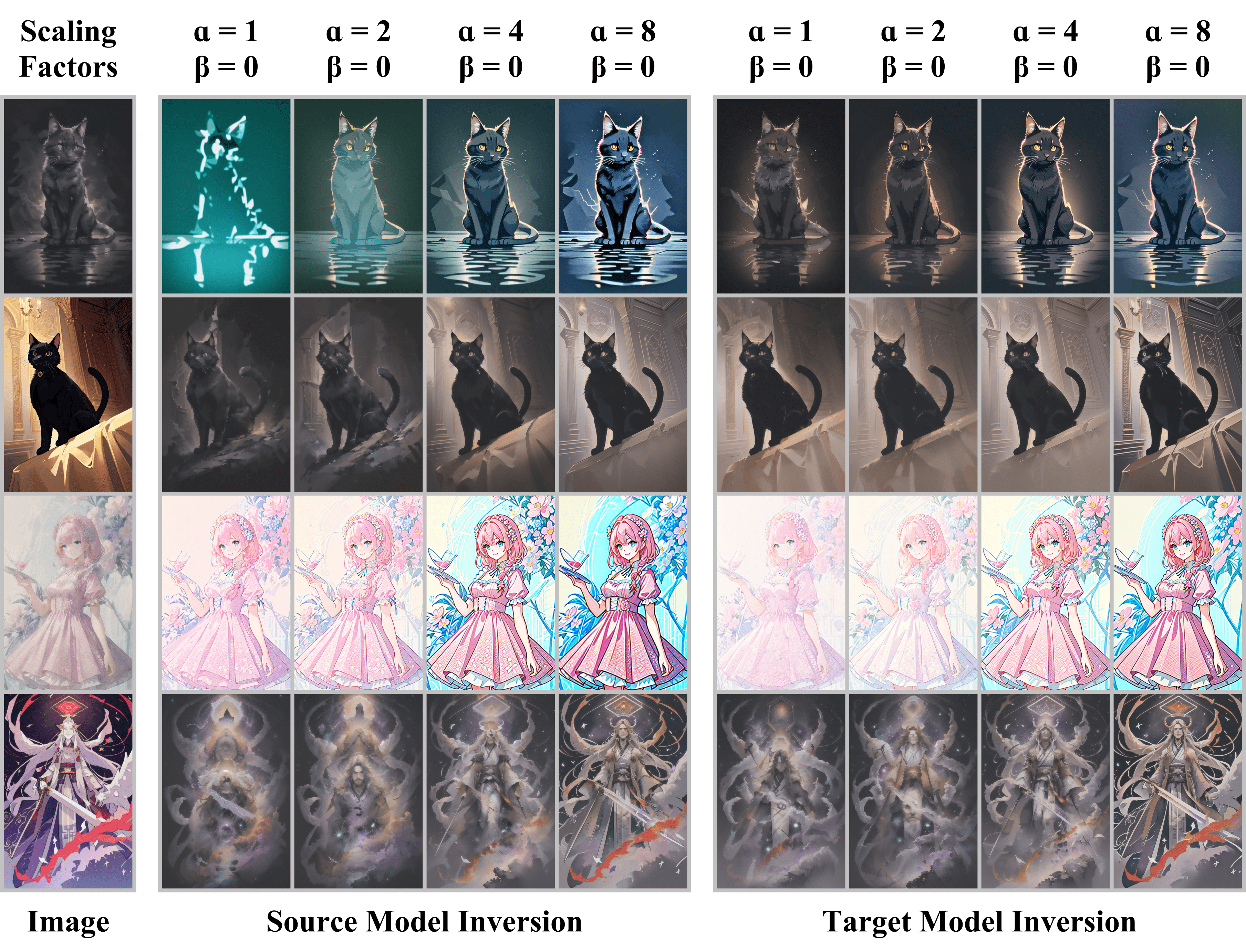}
    \vspace{-8pt}
  \caption{Comparison of performing inversion with source and target DMs. Best view with color.}
\label{fig:sm_inv_compare}
\end{figure*}

One natural variant of Ditail style transfer is to invert a source image with non-original DMs. We perform both qualitative and quantitative investigations on the pros and cons of the variants. In general, performing inversion with the target DM enhances structure preservation but degrades image fidelity and style. Some qualitative results are presented in Fig.~\ref{fig:sm_inv_compare}. The first row demonstrates a rare failure case of inverting with source DM. This happens when the source DM possesses a strong style, yet the target DM fails to denoise the strongly stylized latents properly and leads to a collapse of content and structure. In such cases, inverting with target DMs can give better results. However, this also inevitably degrades the stylizing capability. As shown in the second and third rows, inverting with the target DM preserves the content, but also results in less stylized output.

\begin{table}[h!]
\centering
\scriptsize
\begin{tabular}{c|c|cccc}
\toprule
Inversion DM & Scaling & CLIP $\uparrow$ & DINO $\downarrow$ & \( \text{FID} \downarrow  \) \\
\midrule
\multirow{4}{*}{Photo-realistic} & \( \alpha = 1, \beta=1 \) & 0.2499 & 0.0368 & 127.64 \\
& \( \alpha = 2, \beta=1 \) & 0.2466 & 0.0432 & \textbf{120.65} \\
& \( \alpha = 4, \beta=1 \) & 0.2499 & 0.0385 & 124.60 \\
& \( \alpha = 8, \beta=1 \) & \textbf{0.2505} & \textbf{0.0334} & 146.34 \\
\midrule
\multirow{4}{*}{Target Style} & \( \alpha = 1, \beta=1 \) & 0.2523 & 0.0332 & 130.80 \\
& \( \alpha = 2, \beta=1 \) & 0.2518 & 0.0380 & \textbf{124.34} \\
& \( \alpha = 4, \beta=1 \) & \textbf{0.2525} & 0.0339 & 128.36 \\
& \( \alpha = 8, \beta=1 \) & 0.2516 & \textbf{0.0312} & 149.57 \\
\bottomrule
\end{tabular}
\vspace{4pt}
\caption{Quantitative comparison of inversion with different DMs on real images. The mean CLIP cosine similarity~\cite{hessel2021clipscore, radford2021learning} on source images is 0.2533, and the FID~\cite{heusel2017fid} between source and target images is 154.89. The best-performed metrics are shown in \textbf{bold}.}\looseness=-1
\label{tab:sm_inversion_real}
\end{table}

\begin{figure*}[h!]
  \centering
\includegraphics[width=0.5\textwidth]{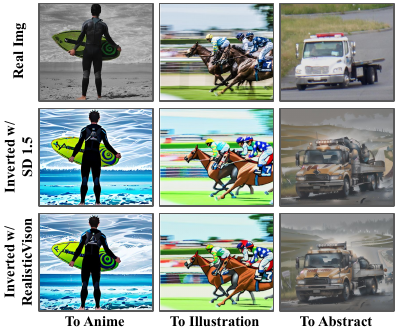}
  \vspace{-4pt}
  \caption{Qualitative comparison of performing inversion with photo-realistic and generic DMs. Zoom in for best view. \looseness=-1}
  \label{fig:inversion_comp}
\end{figure*}

For real images, the quantitative results in Tab.~\ref{tab:sm_inversion_real} demonstrates the trade-off between structure and style. We can see that performing inversion with the target DM tend to enhance structure preservation but degrades the stylization performance. Given that there is no source DM for real-world images, its image distribution can be better approximated by photo-realistic DMs, allowing more flexibility in the structure and thus resulting in better stylized output. We also compare the qualitative results of using photo-realistic (Realistic Vision) or generic (Stable Diffusion 1.5) DMs in Fig.~\ref{fig:inversion_comp} for inversion, where the differences are not significant. By default, we use photo-realistic DMs for real-world images as it is further fine-tuned on this domain.

\begin{figure*}[t!]
  \centering
\includegraphics[width=\textwidth]{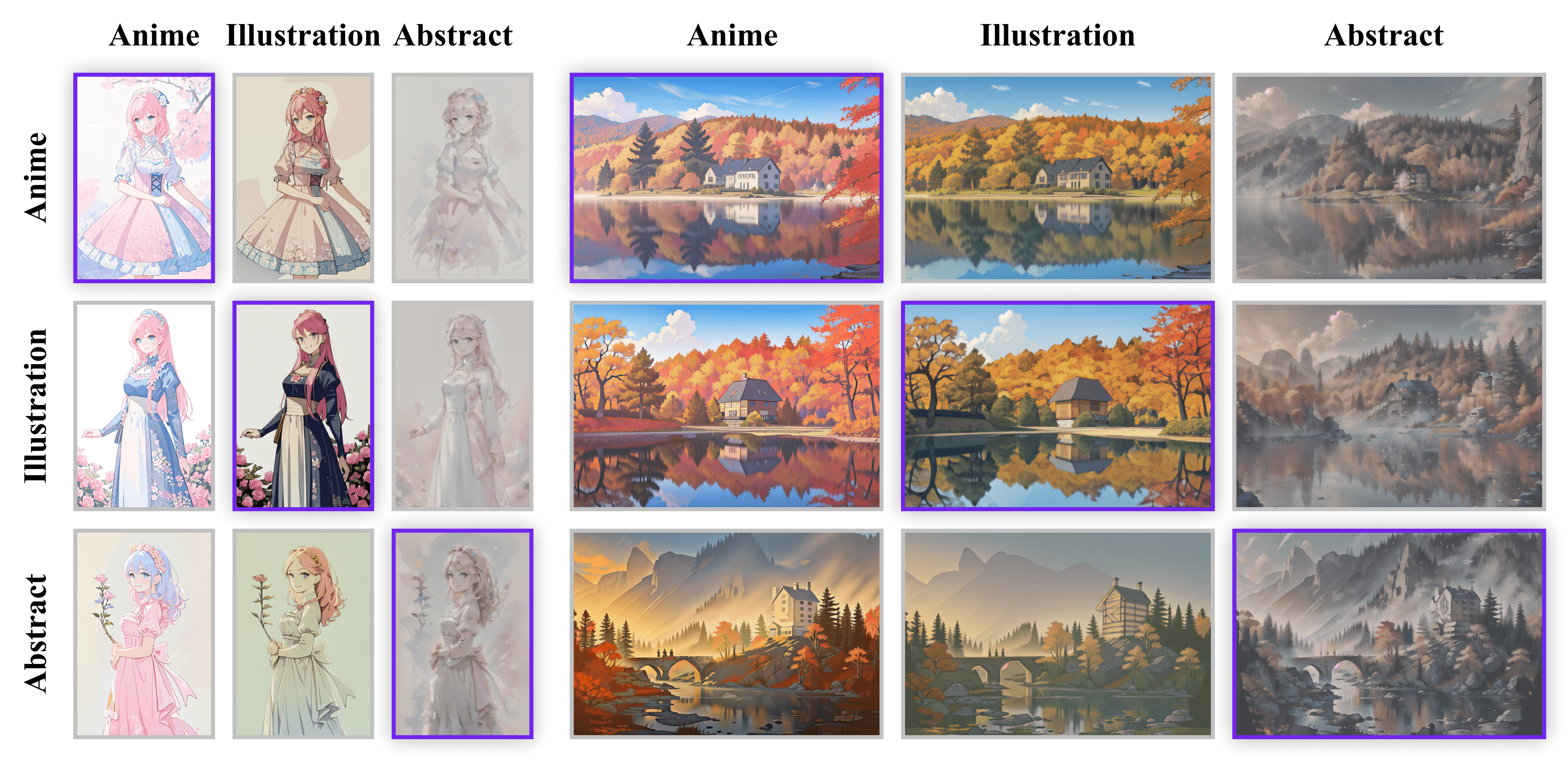}
  \caption{Demonstration of novel image generation across various DMs. Images on the diagonal are generated by the original DMs, whereas the others are the novel results. Ditail enables us to mix content and style information easily on images of all types.}
  \label{fig:sm_novel_gen}
\end{figure*}

\newpage
$ $\newpage
$ $\newpage

\section{More Qualitative Results for Style Transfer}
\label{sec:more_qualitative}
\begin{figure*}[h!]
  \centering
\includegraphics[width=\textwidth]{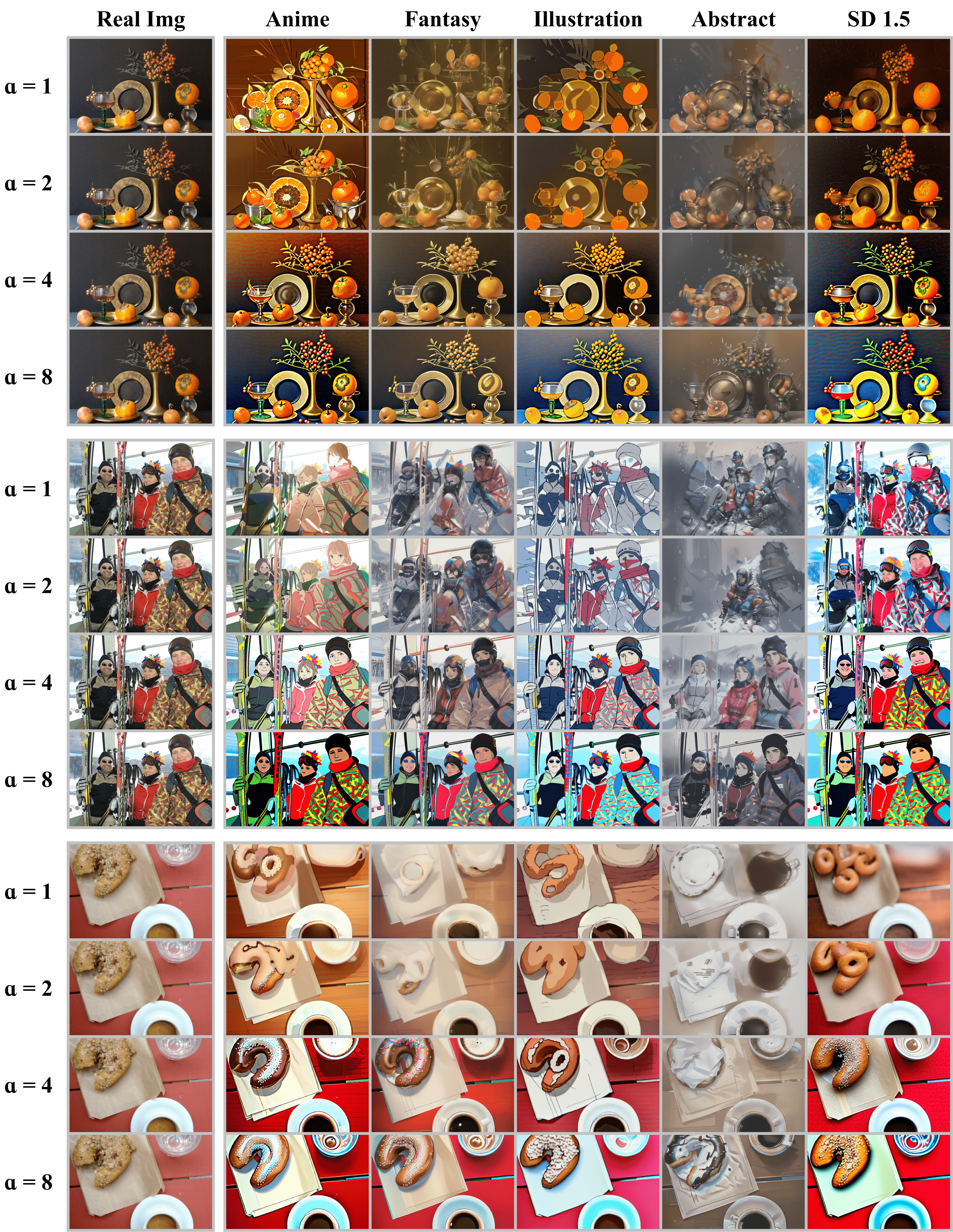}
  \caption{More style transfer results (DM only) on real images. Better view with color.}
  \label{fig:sm_novel_real}
\end{figure*}

\begin{figure*}[h!]
  \centering
\includegraphics[width=\textwidth]{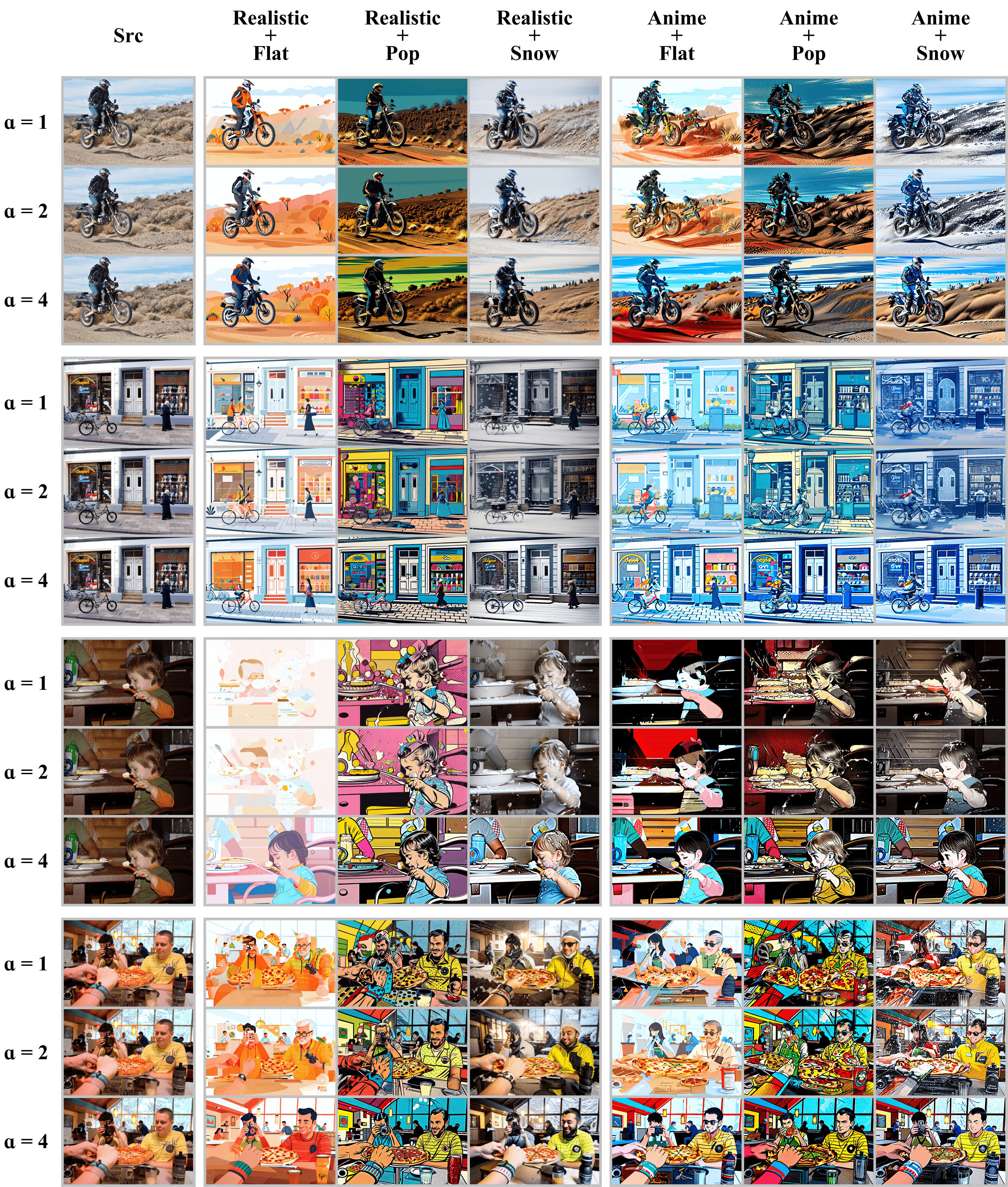}
  \caption{More style transfer results (DM+LoRA) on real images with different scaling factors. During generation, trigger words for each LoRA are appended to the prompt, and we set the scale of LoRA weight to be 0.7, as recommended by LoRA uploaders. We can observe evident transformations in image style and content, while the structure of the source image is largely preserved. We also notice that the effect of the same LoRA on different checkpoints varies, which entails the DM-LoRA compatability problem. Better view with color.}
  \label{fig:sm_lora}
\end{figure*}

We attach more qualitative results produced with Ditail. 
Fig.~\ref{fig:sm_novel_gen} presents the matrix view of cross-transfer across DMs of various styles.
Fig.~\ref{fig:sm_novel_real} and Fig.~\ref{fig:sm_lora} shows real image style transfer results with different model configurations and scaling factors. By taking advantage of the vast collection of DM/LoRA checkpoints, Ditail is able to transfer and mix content/style information in a training-free manner. The presented qualitative results demonstrate its high applicability and generalization ability. In theory, Ditail can generate any subjects of any style, as long as we have the corresponding source images and target DM checkpoints.


\newpage

\newpage


\end{document}